\documentclass[lettersize,journal]{IEEEtran}

\usepackage{amsmath,amsfonts}
\usepackage{array}
\usepackage[caption=false,font=normalsize,labelfont=sf,textfont=sf]{subfig}
\usepackage{textcomp}
\usepackage{stfloats}
\usepackage{url}
\usepackage{verbatim}
\usepackage{graphicx}
\usepackage{cite}
\usepackage{bm}
\usepackage{cutwin}
\usepackage{enumitem}
\usepackage{booktabs}
\usepackage{multirow}
\usepackage{nicefrac}
\usepackage{microtype}
\usepackage[table]{xcolor}
\usepackage{bbding}
\usepackage{pifont}
\usepackage{colortbl}
\usepackage{rotating}
\usepackage{adjustbox}
\usepackage{makecell}
\usepackage{tcolorbox}
\usepackage{algorithm}
\usepackage{algpseudocode}
\usepackage{bbm}

\usepackage{amssymb}
\usepackage{bbm}

\usepackage{amsthm}

\usepackage{mathtools}
\usepackage[pagebackref=false,colorlinks=true, linkcolor = red,citecolor=blue]{hyperref}
\usepackage{cleveref}
\usepackage{wrapfig}
\usepackage{listings}
\usepackage[normalem]{ulem}
\usepackage[utf8]{inputenc}
\usepackage{xurl}

\definecolor{DarkGreen}{HTML}{006400}
\definecolor{DarkRed}{HTML}{8B0000}
\definecolor{diy_pink}{RGB}{255,247,240}
\definecolor{sgreen}{RGB}{30,150,30}
\definecolor{mycolor_blue}{HTML}{E7EFFA}
\definecolor{mycolor_green}{HTML}{E6F8E0}
\definecolor{mycolor_gray}{HTML}{ECECEC}
\definecolor{pearDark}{HTML}{2980B9}
\definecolor{textcolor1}{rgb}{0.25,0.5,0.5}
\definecolor{textcolor2}{rgb}{0.7,0.25,0.25}
\definecolor{linkc}{rgb}{0,0.44,0.74}
\definecolor{eqc}{rgb}{1,0,0}
\definecolor{myy}{RGB}{126,95,0}
\definecolor{mygray}{gray}{.9}
\definecolor{bblue}{RGB}{30,80,120}
\definecolor{mygray1}{gray}{.7}
\definecolor{ggray}{RGB}{127,127,127}
\definecolor{mygreen}{RGB}{93,174,86}
\definecolor{citecolor}{HTML}{229954}
\definecolor{light_green}{HTML}{F5FFFA}
\definecolor{LightCyan}{rgb}{0.88,1,1}
\definecolor{scolor}{RGB}{111,168,220}
\definecolor{hcolor}{RGB}{111,176,81}
\definecolor{ocolor}{RGB}{224,103,102}
\definecolor{wcolor}{RGB}{246,178,107}
\definecolor{dino}{RGB}{227,240,249}

\newcommand{\ie}{{\emph{i.e.}}}
\newcommand{\eg}{{\emph{e.g.}}}

\useunder{\uline}{\ul}{}

\usepackage{tikz}
\usepackage[pagebackref=false,colorlinks=true, linkcolor = red,citecolor=blue]{hyperref}
\definecolor{lime}{HTML}{A6CE39}
\DeclareRobustCommand{\orcidicon}{%
    \begin{tikzpicture}
    \draw[lime, fill=lime] (0,0) 
    circle [radius=0.16] 
    node[white] {{\fontfamily{qag}\selectfont \tiny ID}};    \draw[white, fill=white] (-0.0625,0.095) 
    circle [radius=0.007];    \end{tikzpicture}
    \hspace{-2mm}}
\foreach \x in {A, ..., Z}{%
    \expandafter\xdef\csname orcid\x\endcsname{\noexpand\href{https://orcid.org/\csname orcidauthor\x\endcsname}{\noexpand\orcidicon}}
}

\begin{document}
\bstctlcite{IEEEexample:BSTcontrol}


\title{OmniFysics: Towards Physical Intelligence Evolution via Omni-Modal Signal Processing and Network Optimization}

\author{Minghao Han\orcidC{}, Dingkang Yang\orcidA{}, Yue Jiang\orcidF{}, Yizhou Liu\orcidE{}, Peng Zhai\orcidD{}, Lihua Zhang\orcidC{},~\IEEEmembership{Member, IEEE}

\thanks{Minghao Han, Dingkang Yang, Yue Jiang, Yizhou Liu, Peng Zhai, and  Lihua Zhang are with the College of Intelligent Robotics and Advanced Manufacturing, Fudan University\& Fysics AI, Shanghai 200433, China. (E-mails: \{dkyang20, lihuazhang,\}@fudan.edu.cn); \{mhhan22, jiangyue23, liu\}@m.fudan.edu.cn)}
\thanks{Co-first authors: Minghao Han, Dingkang Yang, and Yue Jiang.}
\thanks{Corresponding authors: Peng Zhai and Lihua Zhang.}
}

\markboth{IEEE Journal of Selected Topics in Signal Processing}%
{Shell \MakeLowercase{\textit{et al.}}: A Sample Article Using IEEEtran.cls for IEEE Journals}


\maketitle

\begin{abstract}
The autonomous evolution of networked AI systems relies heavily on robust environmental perception. However, physical understanding remains brittle in current models because key physical signals are visually ambiguous and sparsely represented in web-scale data. To bridge the gap between data-centric learning and knowledge-based physical rules, we present OmniFysics, a compact omni-modal network that unifies signal processing and understanding across images, audio, video, and text. To enable autonomous optimization and inject explicit physical knowledge, we construct a dynamic physical data engine. Within this engine, FysicsAny acts as an adaptive mechanism that produces physics-grounded supervision by mapping salient objects to verified physical attributes via hierarchical retrieval and physics-law-constrained signal verification. Concurrently, FysicsOmniCap distills web videos utilizing advanced audio-visual cross-modal signal processing, generating high-fidelity data pairs that emphasize dynamic physical cues. We optimize the OmniFysics network through staged multimodal alignment and evolutive instruction tuning, integrating latent-space flow matching for generation and an adaptive intent router for efficient execution. Experiments demonstrate that this evolutive optimization paradigm not only achieves competitive performance on standard multimodal benchmarks but also significantly advances physics-oriented evaluations.
\end{abstract}


\section{Introduction}
\label{sec:intro}

The evolution of Multimodal Large Language Models (MLLMs) is driving a paradigm shift from isolated algorithmic tools toward autonomous and networked AI systems capable of continuous adaptation in complex physical environments~\cite{park2026beyond,chang2025rehazing,yao2025sgnet,li2026cross}. While foundational models ranging from GPT-4o~\cite{gpt4o} to Gemini 3 Pro~\cite{team2023gemini} have established a core foundation with remarkable capabilities in text-image understanding~\cite{bai2025qwen2.5vl, guo2025seed1.5vl, team2025gemma3}, audio-video generation~\cite{xu2025qwen2.5omni,wu2025stepaudio2,team2025klingomni}, and cross-modal reasoning~\cite{xu2025qwen2.5omni, team2025longcatomni,ai2025mingfalshomni}, transitioning these models into evolutive network nodes requires fundamentally rethinking how they process, ground, and optimize real-world multimodal signals.

However, despite these models' increasingly sophisticated performance at the semantic level, they still face a significant knowledge gap at the level of physical perception~\cite{qiu2025phybench}. This lack of physical perception directly leads to current models frequently generating ``physical hallucinations'' that violate causality in generation tasks~\cite{zhu2024sorasurvey} (\eg, liquids flowing backward). In reasoning tasks, they tend to rely on shallow semantic labels rather than physical parameters, severely limiting their generalization capabilities and reliability in the physical world.

The deficit in physical cognition arises not merely from data scarcity, but from the inherent visual ambiguity of physical attributes. Many critical parameters remain latent in static images that rely only on appearance. For example, a steel ball and a painted plastic ball may be visually indistinguishable yet easily categorized via impact sounds or motion cues~\cite{wiesner2025towardsphysicsmodel}. Thus, physical perception often involves cross-modal disambiguation, where complementary sensory inputs help resolve hidden physical properties. Generation can also serve as a probe beyond static classification: synthesizing images that remain consistent with the conditioning context and basic physical regularities can reflect more physics-aware representations~\cite{motamed2025generative}. In OmniFysics, we adopt a latent-space text-to-image generator trained with standard conditional flow matching and leverage verified physical annotations to guide learning. Concretely, we enrich captions by rewriting object mentions into verified physical attributes, strengthening physics–semantics alignment.

Beyond this, we face another critical bottleneck: the extreme scarcity of high-quality physical alignment data, as existing web-scale datasets prioritize semantics over physical grounding. To bridge this gap, we establish a holistic data ecosystem driven by physical heuristics, comprising two core engines tailored to distinct modalities. 
First, we introduce the FysicsAny pipeline for static properties.
This approach employs a five-stage perception-retrieval-verification specialist collaboration mechanism driven by stringent physical constraints, thereby constructing the first large-scale physical property dataset. Complementarily, for dynamic processes, we develop FysicsOmniCap, a framework that distills highly aligned audiovisual samples to enforce temporal physical consistency. Finally, to quantify the emergence of physical intelligence, we propose FysicsEval, a holistic benchmark covering attribute prediction, commonsense reasoning, and cross-modal consistency.

At the architectural level, we present OmniFysics, a new omni-modal large model capable of authentic physical world perception. OmniFysics achieves unified understanding across four modalities (\ie, image, text, audio, and video), and high-fidelity generation for image, text, and audio. To navigate complex cross-modal interactions while balancing inference depth and computational efficiency, we introduce an adaptive dynamic switching mechanism. This mechanism intelligently routes tasks between a lightweight perception pattern and a more advanced physical-visual generation pattern, guided by semantic intent and syntactic priors. Extensive experiments demonstrate that OmniFysics not only achieves significant gains on physical perception benchmarks but also yields substantial, consistent improvements in image-text, audio, and omni-modal understanding tasks. Our work demonstrates injecting physical knowledge into omni-modal models, grounding future embodied intelligence for understanding and interacting with the physical world.
\section{Related Work}
\label{sec:related_work}

\subsection{Omni-Modal Foundation Models}
The research paradigm of AGI is undergoing a profound transformation, shifting from specialized MLLMs to unified, any-to-any Omni-modal LLMs (OLMs). Represented by next-generation models such as GPT-4o~\cite{gpt4o} and Gemini 3~\cite{team2023gemini}, these architectures dismantle the barriers between modalities. Rather than being confined to simple vision-text alignment, they aim to achieve seamless integration and generation of heterogeneous data from different modalities.

Current research on OLMs primarily focuses on extending modalities, optimizing routing mechanisms based on Mixture-of-Experts (MoE) models, and scaling up training, such as Qwen2.5-Omni~\cite{xu2025qwen2.5omni}, Longcat-flash-omni~\cite{team2025longcatomni}, Baichuan-omni-1.5~\cite{li2025baichuan}, and Ming-Flash-Omni~\cite{ai2025mingfalshomni}.
However, current OLMs exhibit systematic deficiencies in physical perception and attribute prediction. This limitation stems from the bias of training data, which predominantly serves semantic understanding while severely lacking high-quality physical perception data capable of explicitly characterizing relationships such as ``impact sound-material density''. Existing methods also lack verifiable physical consistency constraints during training. Therefore, bridging the gap between explicit physical knowledge and implicit neural representations remains a critical bottleneck for current OLMs.

\subsection{Physics AI and Benchmarks}
Recent research targets intuitive physics reasoning in open environments. PhysAgent~\cite{chow2025physbench} augments visual models with physical memory, yet suffers from high inference latency. With the emergence of the world model, Sora~\cite{openai2024sora} and V-JePA~\cite{assran2025vjepa} attempt to simulate the dynamic evolution of the physical world through video generation or latent space prediction. However, despite achieving breakthroughs in generating photorealistic visual appearances, these models face severe challenges regarding physical faithfulness. Most existing works do not actually incorporate intuitive physics constraints or physical laws. They learn pixel-level statistics rather than the physical laws governing the world.

To comprehensively quantify the boundaries of physical cognition in multimodal models, multi-dimensional benchmarks from basic perception to advanced scientific reasoning have been constructed. Early works~\cite{riochet2018intphys, yi2019clevrer} focused on qualitative intuitive-causal physical perception. Recent efforts have shifted towards advanced scientific reasoning and quantitative analysis. QuantiPhy~\cite{puyin2025quantiphy} and ABench-Physics~\cite{zhang2025abench} introduced quantitative physical prediction evaluations. PhysUniBench~\cite{wang2025physunibench}, PHYBench~\cite{qiu2025phybench}, PhysReason~\cite{zhang2025physreason}, and SeePhys~\cite{xiang2025seephys} have introduced physics problems from various educational stages, revealing bottlenecks in handling complex physical laws. Compared to the narrow evaluation scope of PhysToolBench~\cite{zhang2025phystoolbench}, PhysBench~\cite{chow2025physbench} has broadened the domain scope but remains insufficient in evaluation depth. The monotony of its tasks and questions allows models to easily memorize solutions for specific categories of problems. 
In summary, existing evaluation systems still face significant limitations due to the lack of a unified cross-modal mechanism, leaving them unable to comprehensively assess physical property prediction, logical reasoning, and consistency understanding capabilities.

\section{Data Curation Engine}
\label{sec:data_curation}

\begin{figure*}[tp]
    \centering
    \includegraphics[width=0.9\linewidth]{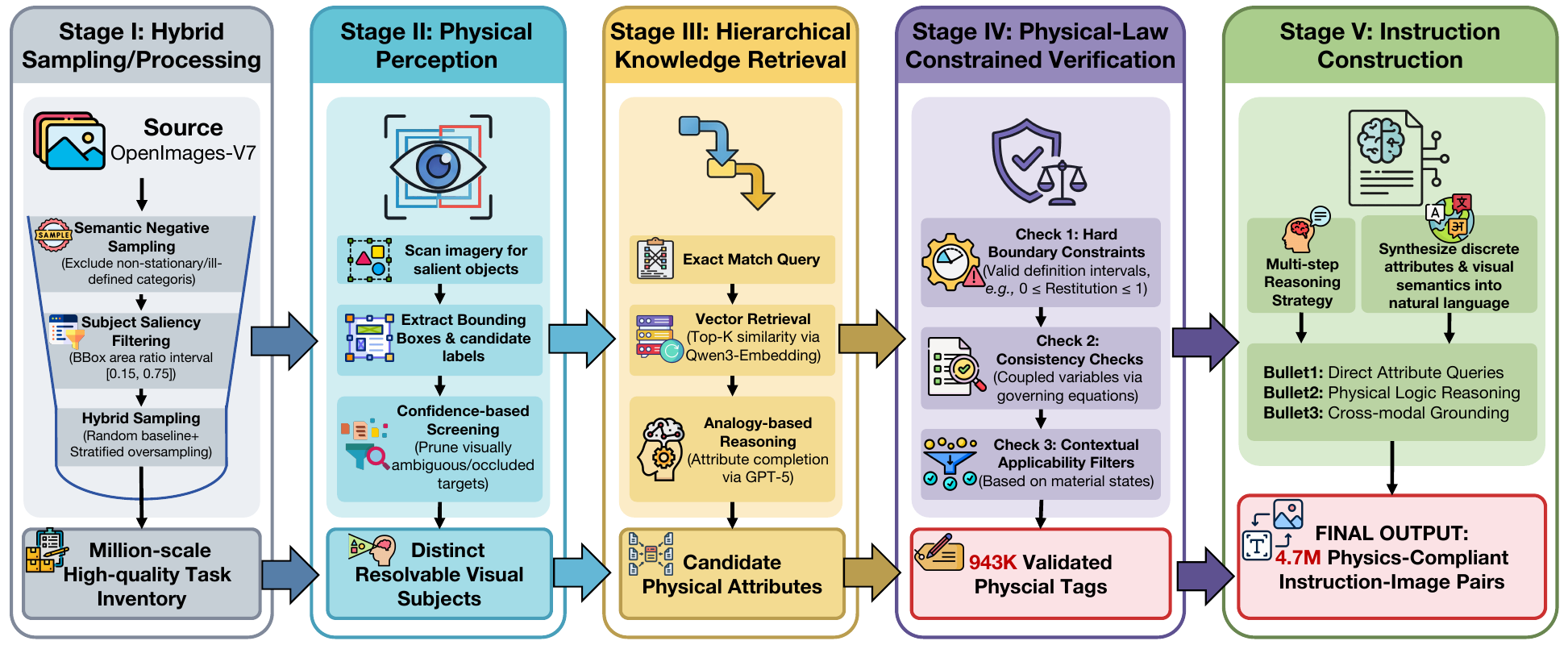}
    \caption{\textbf{FysicsAny Pipeline}. Overview of the pipeline for constructing physics-aware supervision from web-scale data. It integrates object centric perception, hierarchical knowledge retrieval, and physical law constrained verification to generate diverse instruction image pairs for physical attribute supervision.}
    \label{fig:fysics_any}
\end{figure*}

\subsection{Multifaceted Physical Attribute Pipeline}
\label{sec:FysicsAny}

We present FysicsAny, a robust and automated pipeline designed to bridge the gap between visual entities and their intrinsic physical parameters (Figure~\ref{fig:fysics_any}). The pipeline operates through a cascaded five-stage framework. First, we sample object-centric images from OpenImages-V7~\cite{kuznetsova2020openimage} and deploy GPT-5~\cite{openai_introducing_gpt5_2025} for physical perception, applying rigorous bounding-box and confidence-based filtering to isolate salient targets. Next, a Hierarchical Knowledge Retrieval module bridges semantic labels with low-level physical attributes. It queries a curated database of 300 standardized prototypes via a cascaded strategy consisting of exact matching, Qwen3-Embedding~\cite{zhang2025qwen3embedding} retrieval, and GPT-5 analogy reasoning. This process maps objects to an 11-dimensional physical vector ($\mathcal{P}_{vec} \subset \mathbb{R}^{11}$). To strictly eliminate hallucinations, we then enforce physics-law constrained verification, performing hard boundary checks and governing equation consistency. Finally, a Reasoning Specialist (Qwen3-VL-Plus~\cite{bai2025qwen3vltechnicalreport}) synthesizes these verified facts into diverse formats (\eg, Direct Attribute Queries and Physical Logic Reasoning). This rigorous process ultimately transforms raw imagery into a high-quality dataset containing \textbf{943K} physics-law validated tags and \textbf{4.7M} pairs.

\begin{algorithm}[t]
\caption{FysicsOmniCap Data Generation Pipeline}
\label{alg:fysicsomnicap}
\begin{algorithmic}[1]
\State \textbf{Input:} Raw video dataset $\mathcal{D}_{raw}$
\State \textbf{Output:} Physics-aware caption dataset $\mathcal{D}_{final}$
\State \textbf{Params:} Consistency threshold $\tau$, base sample rate $R$
\State \textbf{Modules:} Cross-modal encoder $\Phi$, Visual tool $\mathcal{T}_{vis}$, Audio tool $\mathcal{T}_{aud}$, Physics tool $\mathcal{T}_{phy}$, Brain model $\mathcal{M}$

\State \textbf{Stage 1: Audio-Visual Consistency Filtering}
\State $\mathcal{D}_{clean} \leftarrow \emptyset$
\For{each video $V \in \mathcal{D}_{raw}$}
    \State $K \leftarrow \textsc{SampleKeyframes}(V)$
    \State $S \leftarrow \textsc{AVConsistencyScore}(V, K, \Phi)$
    \If{$S > \tau$}
        \State $\mathcal{D}_{clean} \leftarrow \mathcal{D}_{clean} \cup \{V\}$
    \EndIf
\EndFor

\State \textbf{Stage 2: Multi-Tool Evidence + Caption Synthesis}
\State $\mathcal{D}_{final} \leftarrow \emptyset$
\For{each video $V \in \mathcal{D}_{clean}$}
    \State $(I_{ctx}, I_{act}) \leftarrow \mathcal{T}_{vis}(V)$ \Comment{context \& actions}
    \State $I_{aud} \leftarrow \mathcal{T}_{aud}(V_{audio})$ \Comment{acoustic cues}
    \State $T \leftarrow \textsc{GetAdaptiveTimestamps}(V_{audio}, R)$
    \State $P \leftarrow \textsc{Aggregate}\big(\{\mathcal{T}_{phy}(V[t]) \mid t \in T\}\big)$
    \State $C \leftarrow \mathcal{M}\big(\textsc{ComposePrompt}(I_{ctx}, I_{act}, I_{aud}, P)\big)$
    \State $\mathcal{D}_{final} \leftarrow \mathcal{D}_{final} \cup \{(V, C)\}$
\EndFor

\State \textbf{return} $\mathcal{D}_{final}$
\end{algorithmic}
\end{algorithm}

\subsection{Real-World Physical Attribute Validation}
To further validate the reliability of the FysicsAny pipeline in estimating physical attributes and the authenticity of its generated data, we conducted a set of empirical evaluation experiments in a controlled real-world physical environment. Unlike traditional visual question-answering tasks that rely purely on semantic understanding, this experiment aims to directly quantify the error between the physical parameters predicted by the model and the actual physical world.

We first collected a batch of images encompassing real objects in physical environments. For the key objects in the images, we utilized precise industrial-grade physical measurement instruments to obtain the actual baseline values (Ground Truth, $y_{gt}$) for various physical attributes. For instance, a Universal Testing Machine was used to accurately measure the Young's Modulus and yield stress of soft bodies; a Rheometer was employed to measure the viscosity of fluids; and a Tribometer was used to obtain the static/kinetic friction coefficients of object surfaces. These values, rigorously measured by real machines, serve as the absolute baseline for evaluating model performance.

After acquiring the images and their corresponding real physical attributes, we fed the same visual inputs into the FysicsAny pipeline and the current state-of-the-art large language model, GPT-5~\cite{openai_introducing_gpt5_2025}, and asked both to independently estimate the objects' physical parameters ($y_{pred}$). To rigorously evaluate the numerical predictions, we adopt the \textit{Mean Relative Accuracy} ($\mathcal{MRA}$) metric~\cite{yang2025thinking}. Rather than relying on a singular, rigid tolerance bound, $\mathcal{MRA}$ provides a comprehensive assessment by aggregating the accuracy across a discrete spectrum of confidence thresholds, denoted as $\mathcal{C} = \{0.5, 0.55, \dots, 0.95\}$. Specifically, for a given ground truth value $y$ and the model's prediction $\hat{y}$, the prediction is considered correct under a specific threshold $\theta \in \mathcal{C}$ if its relative error, $\frac{|\hat{y} - y|}{y}$, strictly falls below the tolerance margin of $1 - \theta$. By averaging these binary accuracy results over the 10 distinct threshold levels, the overall metric is formulated as:

\begin{equation}
    \mathcal{MRA} = \frac{1}{10} \sum_{\theta \in \mathcal{C}} \mathbbm{1} \left( \frac{|y_{pred} - y_{gt}|}{y_{gt}} < 1 - \theta \right).
    \label{eq:mra}
\end{equation}

By evaluating the relative error across this varying range of strictness, $\mathcal{MRA}$ effectively overcomes the evaluation bias introduced by a single threshold, thereby serving as a more robust, discriminative, and reliable indicator of the true numerical alignment between the predictions and the physical ground truth. This metric intuitively reflects the relative degree to which the model's predicted values deviate from the real physical values. A higher $\mathcal{MRA}$ indicates that the estimation of physical attributes is closer to the real physical world.

\begin{figure*}[t]
    \centering
    \includegraphics[width=1\linewidth]{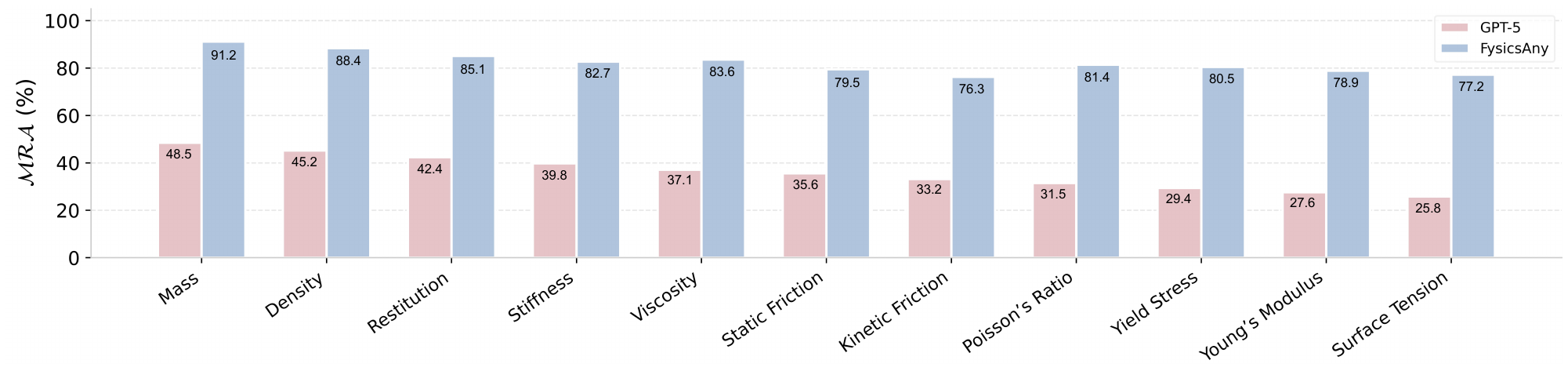}
    \caption{\textbf{Quantitative comparison of Mean Relative Accuracy ($\mathcal{MRA}$) in real-world physical attribute estimation.} The chart evaluates the predictive performance of GPT-5 and the proposed FysicsAny pipeline across 11 diverse physical properties against instrument-measured ground truths. Higher $\mathcal{MRA}$ percentages indicate closer alignment with actual real-world physical values. FysicsAny consistently and significantly outperforms GPT-5 across all parameters, demonstrating its superior capability to overcome visual ambiguity and accurately estimate complex physical properties.}
    \label{fig:fysicsany_bar}
\end{figure*}

The comparative experimental results are illustrated in Figure~\ref{fig:fysicsany_bar}. Although GPT-5 possesses strong generalization and semantic representation capabilities, its lack of explicit physical law constraints makes it highly susceptible to physical hallucinations when estimating specific physical values (e.g., density, stiffness, restitution), resulting in a significantly lower $\mathcal{MRA}$ score. In contrast, FysicsAny exhibits exceptionally high measurement accuracy and robustness. Thanks to the hierarchical knowledge retrieval and Physical-Law Constrained Verification mechanisms within its architecture, FysicsAny can effectively eliminate visual ambiguity and accurately anchor visual features to reasonable physical parameter intervals. This comparative experiment not only provides strong evidence for the high fidelity of FysicsAny in extracting physical attributes within real-world environments but also further validates the absolute quality and training value of the million-scale physical attribute dataset constructed based on this pipeline.

\subsection{Dynamic Omni Alignment Engine}
\label{sec:FysicsOmniCap}
While FysicsAny grounds static properties, the physical world inherently relies on temporal dynamics and cross-modal causality. To this end, we propose FysicsOmniCap, a two-stage engine that constructs large-scale physics-aware video-audio instruction datasets. 

\vspace{0.5pt}\noindent\textbf{Stage I: Audio-Visual Dynamic Purification.}\quad 
To distill high-information clips from noisy web data (\ie, VGGSound~\cite{chen2020vggsound}), we employ a hybrid keyframe sampling strategy: primarily uniform sampling, supplemented by interpolation at audio transient peaks. Using the ImageBind~\cite{girdhar2023imagebind}, we compute segment-level cosine similarities between aligned audio windows and visual keyframes. A loose threshold is then applied to prune irrational samples, preserving basic causal consistency without over-discarding valid data.

\vspace{0.5pt}\noindent\textbf{Stage II: Brain-Tool Collaboration.}\quad 
The purified data is processed via a centralized orchestration framework. A Central Brain (Gemini-2.5-Flash~\cite{comanici2025gemini2.5}) coordinates three specialized Perception Tools: 
(1) \textbf{General Visual Analyzer} (Qwen2.5-VL-72B~\cite{bai2025qwen2.5vl}) extracts macroscopic scene semantics (actions and contexts). 
(2) \textbf{Audio Detail Analyzer} (Qwen2-Audio~\cite{chu2024qwen2audio}) captures physical acoustic cues, such as material properties or structural tendencies. 
(3) \textbf{Physics-Perception Specialist} (Qwen2.5-VL-7B~\cite{bai2025qwen2.5vl}, fine-tuned on FysicsAny) utilizes adaptive sampling by increasing frame density at audio transients to output frame-level structured physical tags.

\vspace{0.5pt}\noindent\textbf{Synthesis of Physical Causal Chains.}\quad 
Finally, the Central Brain spatially and temporally aligns these multi-source signals to synthesize deep \textit{Physical Causal Chains}. Transcending surface-level captioning, it generates physics-aware narratives; for example, inferring that a highly rigid object exhibits no plastic deformation upon impact, rebounds, and synchronously excites a high-frequency, slowly decaying transient sound. Ultimately, this engine yields \textbf{872K} high-fidelity video--instruction pairs, serving as crucial supervisory signals for temporal and cross-modal physical reasoning.

\subsection{FysicsEval: Physical Intelligence Benchmark}
\label{sec:FysicsEval}

Existing physics benchmarks predominantly target theoretical problem-solving or qualitative scenario analysis, as detailed in Table~\ref{tab:related-bmk}. They typically assess only intuitive physics or university-level question proficiency, which is insufficient for the next generation of generalist MLLMs and world models designed to interact with physical reality.  To enable a comprehensive assessment of MLLMs’ perception, reasoning, and understanding in the physical world, we present FysicsEval, a holistic benchmark for multimodal physical intelligence. Diverging from prior evaluation systems limited to qualitative intuition or isolated domains, FysicsEval offers a holistic, multi-granular assessment emphasizing three core competencies: (i) quantitative prediction of physical attributes, (ii) interpretable reasoning grounded in physical laws, and (iii) cross-modal understanding with physical consistency. This design rigorously probes whether models can transcend superficial semantic alignment to achieve robust physical perception via cross-modal evidence. FysicsEval establishes a unified standard for assessing Physical AI. 

\textbf{Composition and Diversity.} FysicsEval comprises 3,854 samples and 3,781 images captured from real-world scenarios, covering three tasks: physical attribute prediction, physical reasoning, and physical consistency understanding. To mitigate rote memorization, we diversify the query formats into numerical prediction, open-ended questions, and Multiple-Choice Questions (MCQ). All queries are stratified into three difficulty levels to accommodate models at varying stages of capability. FysicsEval encompasses three representative physical states: rigid bodies, soft bodies, and fluids, thereby spanning a spectrum from commonsense physics to complex dynamic phenomena and engineering attributes. Furthermore, we define an attribute space in FysicsEval, consisting of 11 critical categories centered on parameters essential for physical simulation and engineering modeling, including: \textit{stiffness}, \textit{density}, \textit{mass}, \textit{coefficient of static friction}, \textit{coefficient of kinetic friction}, \textit{restitution}, \textit{Young’s modulus}, \textit{Poisson’s ratio}, \textit{viscosity}, \textit{surface tension}, and \textit{yield stress}.
These attributes include both latent variables inferable from multimodal cues (\eg, density, Young's modulus) and dynamic parameters highly dependent on interaction processes (\eg, friction coefficients, restitution). This design explicitly probes whether models can transcend superficial semantic alignment to achieve robust physical perception via cross-modal evidence. Consequently, FysicsEval stands as a unique physical benchmark capable of comprehensively evaluating MLLMs across three dimensions: attribute precision, reasoning depth, and physical consistency.

\textbf{Task Taxonomy.} To provide a multi-granular assessment of Physical AI, FysicsEval establishes a unified evaluation framework encompassing three complementary capabilities:
\begin{itemize}[leftmargin=*, noitemsep, topsep=2pt]
    \item \textbf{Perception and Prediction of Physical Attribute.} Given visual inputs of real-world scenarios and textual conditions, models are required to produce precise quantitative estimates of target attributes. The goal is to measure the ability to map observable phenomena to underlying parameters, evaluating whether models can recover implicit physical quantities by leveraging complementary multimodal evidence rather than relying on category priors.
    \item \textbf{Explainable Physical Reasoning.} This task targets the derivation of physical laws and causal chains. Given initial conditions, constraints, and observational cues, models are required to generate conclusions that adhere to physical laws. This measures systematic reasoning and explainability regarding core principles such as conservation laws, force analysis, material response, and fluid dynamics.
    \item \textbf{Cross-modal Physical Consistent Understanding.} Focusing on consistency within the physical world, this task requires models to integrate multimodal attributes and commonsense laws to identify physical hallucinations that violate causality or common sense, thereby assessing the robustness of real-world understanding under complex conditions.
\end{itemize}

\begin{table*}[t]
\caption{\textbf{Comparison of Phsical Benchmarks.} MCQ / OE denote multiple-choice / open-ended questions. \textit{Pre} denotes numerical prediction.}
\renewcommand{\arraystretch}{1.2}%
\resizebox{\textwidth}{!}{%
\begin{tabular}{lccccccc}
\hline
\textbf{Benchmark}                                         & \textbf{Questions} & \textbf{\begin{tabular}[c]{@{}c@{}}Question\\ Type\end{tabular}} & \textbf{\begin{tabular}[c]{@{}c@{}}Physical Percption\\ \& Prediction\end{tabular}} & \textbf{\begin{tabular}[c]{@{}c@{}}Physical\\ Understanding\end{tabular}} & \textbf{\begin{tabular}[c]{@{}c@{}}Physical\\ Reasoning\end{tabular}} & \textbf{\begin{tabular}[c]{@{}c@{}}Real-World\\ Image\end{tabular}} & \textbf{\begin{tabular}[c]{@{}c@{}}Physical\\ Metrics\end{tabular}} \\ \hline
PhysBench                                                  & 10,002             & MCQ                                                              & \textcolor{red}{\ding{55}}                                                                          & \textcolor{green}{\ding{51}}                                                                & \textcolor{red}{\ding{55}}                                                            & \textcolor{green}{\ding{51}}                                                          & \textcolor{red}{\ding{55}}                                                          \\
PhysUniBench                                               & 3,304              & MCQ / OE                                                           & \textcolor{red}{\ding{55}}                                                                          & \textcolor{green}{\ding{51}}                                                                & \textcolor{green}{\ding{51}}                                                            & \textcolor{red}{\ding{55}}                                                          & \textcolor{green}{\ding{51}}                                                          \\
PAI-Bench-Und.                                             & 2,808              & MCQ                                                              & \textcolor{red}{\ding{55}}                                                                          & \textcolor{green}{\ding{51}}                                                                & \textcolor{red}{\ding{55}}                                                            & \textcolor{green}{\ding{51}}                                                          & \textcolor{red}{\ding{55}}                                                          \\
PhysToolBench                                              & 1000               & MCQ                                                              & \textcolor{red}{\ding{55}}                                                                          & \textcolor{green}{\ding{51}}                                                                & \textcolor{red}{\ding{55}}                                                            & \textcolor{green}{\ding{51}}                                                          & \textcolor{red}{\ding{55}}                                                          \\
SeePhys                                                    & 2,000              & OE                                                               & \textcolor{red}{\ding{55}}                                                                          & \textcolor{green}{\ding{51}}                                                                & \textcolor{green}{\ding{51}}                                                            & \textcolor{red}{\ding{55}}                                                          & \textcolor{red}{\ding{55}}                                                          \\
PHYBench                                                   & 500                & OE                                                               & \textcolor{red}{\ding{55}}                                                                          & \textcolor{red}{\ding{55}}                                                                & \textcolor{green}{\ding{51}}                                                            & \textcolor{red}{\ding{55}} (only text)                                              & \textcolor{green}{\ding{51}}                                                          \\
PhysReason                                                 & 1,200              & OE                                                               & \textcolor{red}{\ding{55}}                                                                          & \textcolor{green}{\ding{51}}                                                                & \textcolor{green}{\ding{51}}                                                            & \textcolor{red}{\ding{55}} (only text)                                              & \textcolor{green}{\ding{51}}                                                          \\
QuantiPhy                                                  & 3,300              & OE                                                               & \textcolor{green}{\ding{51}}                                                                          & \textcolor{red}{\ding{55}}                                                                & \textcolor{red}{\ding{55}}                                                            & \textcolor{green}{\ding{51}}                                                          & \textcolor{red}{\ding{55}}                                                          \\
\rowcolor{dino}   \textbf{FysicsEval (Ours)} & 3,854               & MCQ / OE / \textit{Pre.}                                                      & \textcolor{green}{\ding{51}}                                                                          & \textcolor{green}{\ding{51}}                                                                & \textcolor{green}{\ding{51}}                                                            & \textcolor{green}{\ding{51}}                                                          & \textcolor{green}{\ding{51}}                                                          \\ \hline
\end{tabular}
\label{tab:related-bmk}
}
\end{table*}

\textbf{Evaluation Metrics.} FysicsEval supports unified evaluation tailored to distinct output formats. We employ $\mathcal{MRA}$ (as shown in Eq.~\ref{eq:mra}) for physical attribute prediction, and \textit{accuracy} for consistency understanding presented as MCQ. For the open-ended reasoning task, we utilize an advanced LLM-based judging protocol that evaluates responses across 6 dimensions: semantic consistency, physical parameter precision, physical causal validity, physical mechanism identification, reasoning chain completeness, and quantitative-qualitative alignment. GPT-5 is employed as the evaluator under a standardized scoring protocol. To ensure reliability, we validate the Pearson correlation coefficient~\cite{benesty2009pearson} between GPT-5 and three physics experts to quantify agreement in scoring behavior. The correlation remains consistently high ($r>0.9$), indicating strong concordance between the LLM evaluator and human experts.
\section{Methodology}
\label{methodology}

\subsection{Model Architecture}
\vspace{0.5pt}\noindent\textbf{Visual Encoder.}\quad
We employ SigLIP2-So400m~\cite{tschannen2025siglip2} as the visual encoder. Inspired by Qwen2.5-VL~\cite{bai2025qwen2.5vl}, we design the adapter to achieve 4$\times$ spatial downsampling via $2 \times 2$ pooling. Subsequently, a two-layer MLP projects these condensed features into the LLM's embedding space.

\vspace{0.5pt}\noindent\textbf{Audio Encoder.}\quad
For audio inputs, we resample waveforms to 16kHz and compute 128-channel Mel-spectrograms using a 25 ms window and a 10 ms stride. Whisper-large-v3~\cite{radford2023robustwhisper} is employed as the audio backbone to extract continuous representations. Subsequently, an adapter with 1D average pooling and a linear layer projects the features into the LLM embedding space.

\vspace{0.5pt}\noindent\textbf{Backbone \& Positional Encoding.}\quad
Qwen2.5-3B~\cite{qwen2025qwen25technicalreport} was employed as the LLM. To enable precise spatiotemporal alignment, we adopt Temporal Multimodal Rotary Position Embedding (TMRoPE)~\cite{xu2025qwen2.5omni}, which factorizes positional information into temporal (quantized at a 40ms interval resolution), height, and width components, while interleaving video and audio representations in 2-second chunks.

\vspace{0.5pt}\noindent\textbf{Audio Generation.}\quad
We employ WavTokenizer~\cite{ji2025wavtokenizerefficientacousticdiscrete} for audio generation. Its single-codebook design (4,096 vocabulary) and high compression (40 tokens/s) facilitate efficient autoregressive modeling. Our TTS module, named SpokenVoxer, initialized from Qwen2.5-0.5B~\cite{qwen2025qwen25technicalreport}, predicts discrete audio codes from multimodal context, which are then decoded into waveforms.

\vspace{0.5pt}\noindent\textbf{Image Generation.}\quad
We use a DiT~\cite{peebles2023scalabledit} flow head, trained via Flow Matching in the WAN 2.2 VAE~\cite{wan2025} latent space. It processes concatenated text and patchified image latents using $N$ Adaptive Layer Norm (AdaLN)-modulated Transformer blocks. The model predicts the velocity field $\mathbf{v}_t = \mathrm{d}\mathbf{x}_t/\mathrm{d}t$, enabling efficient sampling via Optimal Transport paths.

\begin{figure*}[tp]
    \centering
    \includegraphics[width=0.9\linewidth]{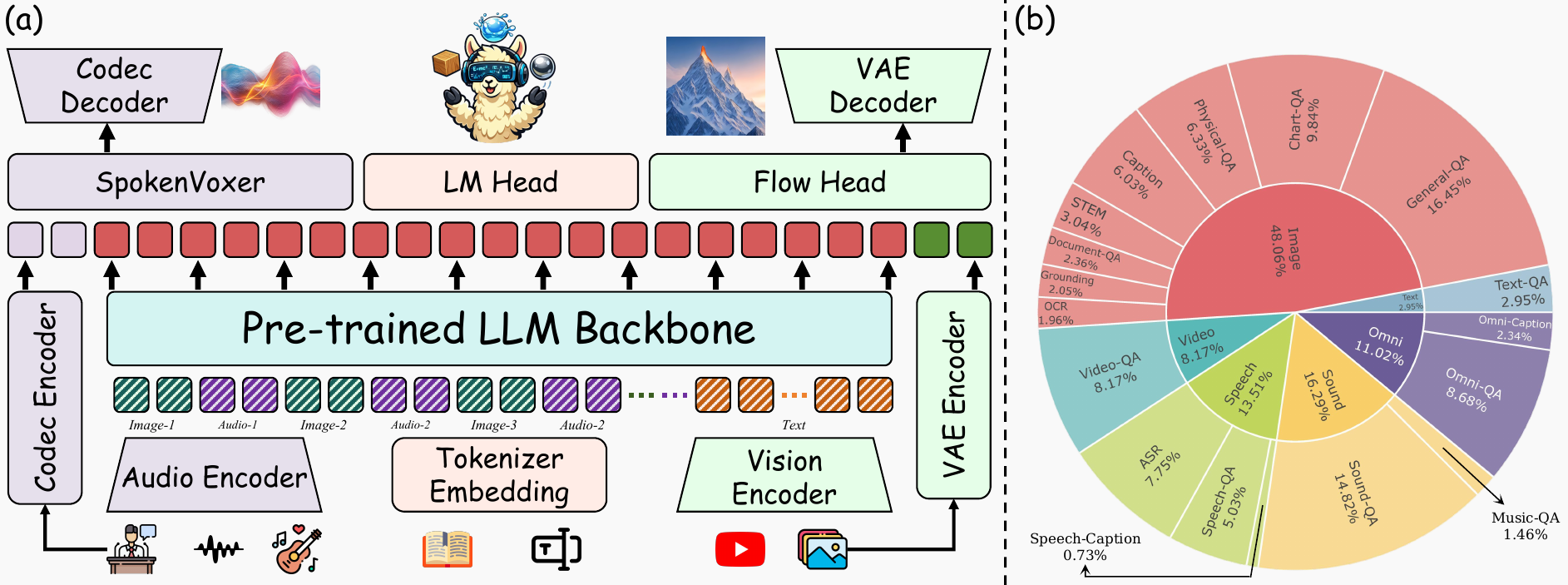}
    \caption{\textbf{Overview of OmniFysics and training data distribution.} \textbf{(a) Model architecture.} The model employs Temporal Multimodal Rotary Position Embedding to process interleaved sequences of images, audio, and text. For understanding, the Vision and Audio Encoders extract features to feed the LLM backbone. For the generation task, the Codec and VAE Encoder are utilized to assist the SpokenVoxer and Flow Head in synthesizing audio and imagery. \textbf{(b) Data distribution for Omni-modal Joint Training.} The pie chart illustrates the modal proportions specific to this training stage: image (48\%), sound (16\%), speech (14\%), omni (11\%), video (8\%), and text-only (3\%).}
    \label{fig:framework_and_data}
\end{figure*}

\subsection{Training Strategy}
\begin{figure*}[t]
    \centering
    \includegraphics[width=0.98\linewidth]{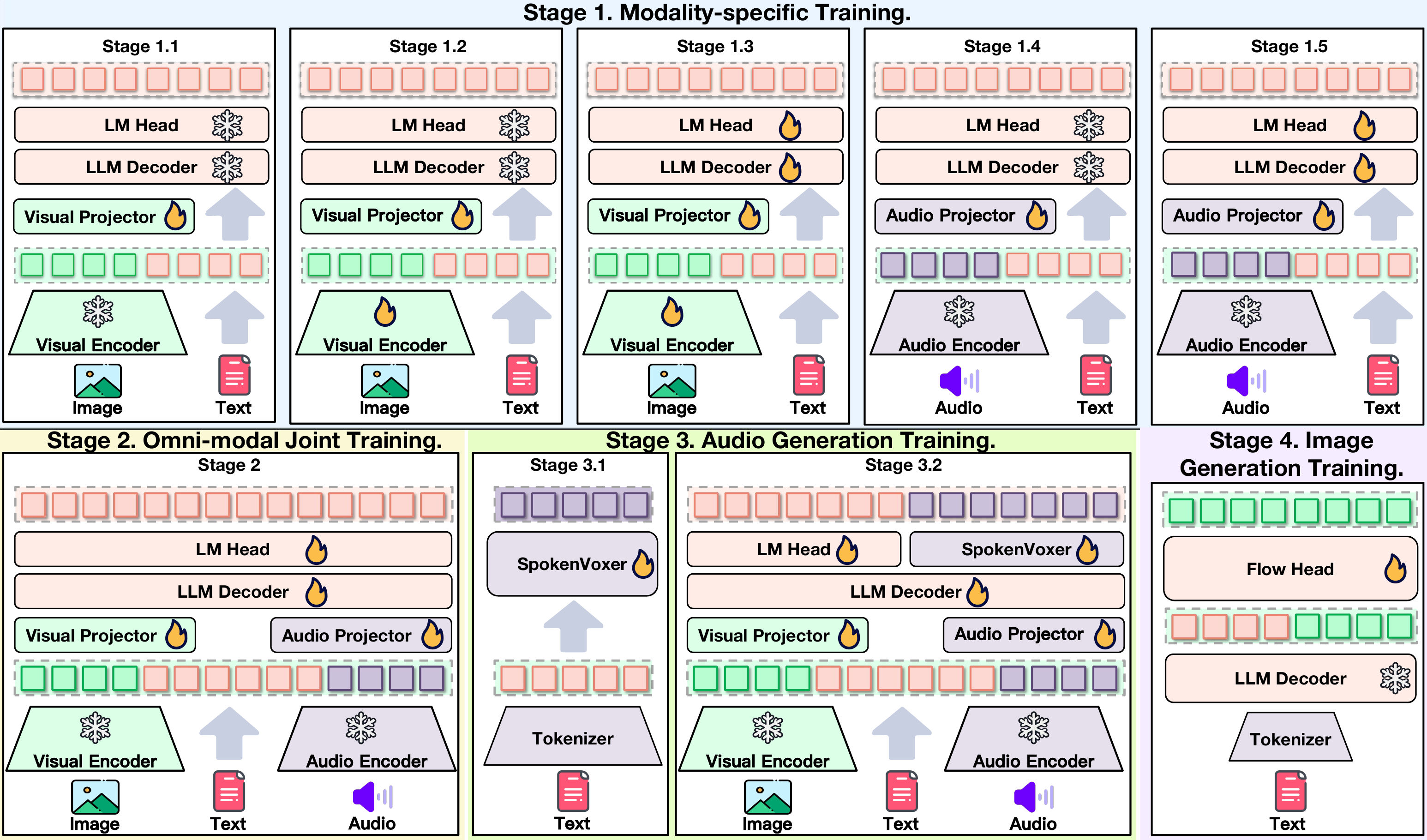}
    \caption{\textbf{Training pipeline of OmniFysics.}
    We employ a four-stage training strategy for OmniFysics to progressively enhance its omni-modal perception and physical understanding, including speech and text-to-image generation.}
    \label{fig:training_stage}
\end{figure*}

To achieve comprehensive multimodal alignment and generation, we adopt a four-stage strategy: (i) modality pre-training, (ii) omni-modal joint training, (iii) audio generation, and (iv) image generation integration (see Figure~\ref{fig:training_stage}).
Formally, let $\mathcal{M}_{\theta}$ denote the multimodal LLM. Given an instruction example $\xi=(x^{(t)},x^{(v)},x^{(a)},y)$, we pack all inputs into a unified conditioning sequence $H$:
\begin{small} 
\begin{equation}
H=\mathrm{Pack}\!\left(x^{(t)},\,\mathcal{A}_v(\mathcal{E}_v(x^{(v)})),\,\mathcal{A}_a(\mathcal{E}_a(x^{(a)}))\right),
\end{equation}
\end{small}%
where $\mathrm{Pack}(\cdot)$ applies the TMRoPE and time-interleaving schemes. This unified representation $H$ serves as the conditioning context for all subsequent generation tasks.

\subsubsection{Modality-Specific Training}
We initialize our framework with the pre-trained Qwen2.5-3B~\cite{qwen2025qwen25technicalreport}. In the first stage, we independently bolster the model's proficiency in visual and audio understanding using unimodal datasets. We apply the standard three-stage paradigm for vision and a lightweight audio alignment strategy that trains the audio adapter and the LLM while keeping the audio encoder frozen, optimizing the next-token prediction objective:
\begin{small}%
\begin{equation}
\mathcal{L}_{\mathrm{LM}}
=
-\mathbb{E}\Big[\sum_{i=1}^{|y|}\log p_{\theta}(y_i \mid y_{<i}, H)\Big],
\label{eq:loss_lm}
\end{equation}
\end{small}%
where $H$ denotes the available unimodal input (\textit{i.e.}, image or audio) with the instruction prompt.

\subsubsection{Omni-modal Joint Training}
\label{omni_joint_train}
In this phase, we unfreeze the LLM and projectors for joint optimization. Data quality and diversity are paramount for effective multimodal alignment. We curate a comprehensive instruction-tuning dataset of 37M samples. Instead of relying on empirical manual settings for the data mixture, the proportions (\eg, 48\% image, 16\% sound, 14\% speech, as shown in Figure \ref{fig:framework_and_data}b) are emergent results from an Information Entropy Maximization strategy.  To prevent the model from being biased towards naturally high-resource modalities (\eg, text) and to ensure balanced activation across physical and semantic representations, we formulate the mixture selection as an optimization problem. Specifically, we evaluate the information density of each sample by computing the sum of Inverse Document Frequencies (IDF) for its extracted N-gram concepts. After filtering out low-entropy, redundant samples, we optimize the mixture weights $w$ to maximize the Shannon entropy of the global feature activation distribution, formulated as $\arg\max_{w} (-\sum_{i} p_i(w) \log p_i(w))$, where $p_i(w)$ denotes the activation probability of the $i$-th feature under the mixture weights. We then optimize Eq. \ref{eq:loss_lm} on this corpus. The specific composition of each data component is detailed as follows.

\vspace{0.5pt}\noindent\textbf{Visual Understanding.}\quad
Static images form the foundation of fine-grained perception. This subset aggregates 163 public datasets spanning eight tasks, totaling 17.9M samples. Notably, to bridge the physical perception gap, we incorporate a Physical-QA subset generated via our FysicsAny engine. Unlike traditional VQA which queries shallow semantics, this data focuses on reasoning about latent physical properties (\eg, material identification, mass estimation, force analysis). These physically constrained samples compel the model to transcend surface-level pixels, effectively mitigating physical hallucinations and internalizing a generalizable world model. Simultaneously, to compensate for the temporal deficiency of static images, we incorporate Video-QA data. This subset focuses on modeling temporal dynamics, motion variations, and event logic.

\vspace{0.5pt}\noindent\textbf{Audio Perception.}\quad
To establish comprehensive auditory capabilities, we integrate 13.5\% speech data (ASR and Speech-QA) for linguistic processing, and 16.3\% non-speech data (Music-QA and Sound-QA) for general acoustic signal understanding.

\vspace{0.5pt}\noindent\textbf{Omni-Modal Alignment.}\quad
Crucially, we introduce this subset as the ``connective tissue'' for multimodal perception, comprising two key components:
\textbf{(i) Omni-QA:} To simulate authentic ``see-and-hear'' interactions, we curate samples from high-quality Image-QA datasets and convert textual questions into audio via open-source TTS (\ie, CosyVoice2~\cite{du2024cosyvoice2}). This enforces a direct mapping between visual content and spoken instructions.
\textbf{(ii) Omni-Caption:} Generated via our FysicsOmniCap engine, this subset utilizes video content with rich audio to provide strictly aligned audiovisual descriptions. This ensures the model captures deep physical causal consistency between visual dynamics and auditory cues.

\subsubsection{Audio Generation Training}
To equip the model with high-fidelity speech generation, we adopt a two-stage TTS recipe. \textbf{Stage I (Foundational Alignment)}: We train our SpokenVoxer on 10M samples to establish a robust text-to-acoustic mapping, freezing other components to prevent interference. \textbf{Stage II (Emotion-Context Joint Training)}: We jointly optimize our SpokenVoxer and LLM using 1.5M Speech-to-Speech (S2S) samples, synthesized with emotion tags via Gemini-2.5-flash~\cite{comanici2025gemini2.5}. To enable emotion-aware generation, we condition generation on the query speech encoding $q^{(a)}$ and target text $y^{(t)}$. We mix 10\% data from Section~\ref{omni_joint_train} to prevent forgetting. The objective maximizes the likelihood of audio tokens $c_{1:T}$ conditioned on context $H$ (containing history and query audio) and text $y^{(t)}$:
\begin{small}
\begin{equation}
\mathcal{L}_{\mathrm{audio\_gen}}
=
-\mathbb{E}\Big[\sum_{t=1}^{T}\log p_{\psi}(c_t \mid c_{<t}, \underbrace{H}_{\text{context}}, \underbrace{y^{(t)}}_{\text{text}})\Big].
\label{eq:loss_audio_gen}
\end{equation}
\end{small}%

\subsubsection{Image Generation Training}
Building upon latent-space Flow Matching~\cite{lipman2022flow} and the DiT flow-head design~\cite{peebles2023scalabledit}, we adopt a three-stage curriculum to progressively expand generative capacity.

\vspace{0.5pt}\noindent\textbf{Stage I: Low-Resolution Pre-training.}\\
We train a 0.9B-parameter DiT flow head from scratch with 75.9M text-to-image (T2I) pairs at $512\times512$ resolution. Conditioned on LLM hidden states, this stage establishes stable global semantic alignment.

\vspace{0.5pt}\noindent\textbf{Stage II: High-Resolution Adaptation.}\quad
We then upsample to $1024\times1024$ and finetune with 10.3M T2I samples. This phase prioritizes high-frequency detail refinement and structural consistency while preserving semantics.

\vspace{0.5pt}\noindent\textbf{Stage III: Physics-Grounded Refinement.}\quad
Finally, we focus on enforcing physical fidelity and precise semantic adherence at $1024\times1024$. In this stage, we curate a high-quality T2I mixture comprising 12.4M aesthetic-filtered samples and $\sim$200K physics-enhanced samples from FysicsAny (Sec.~\ref{sec:FysicsAny}). Unlike previous stages, we apply stricter data quality controls (aspect ratio $\le 2{:}1$, aesthetic score $>6$~\cite{li2024laionsgenhancedlargescaledataset}) and leverage dense captioning via Qwen3-VL-Plus~\cite{bai2025qwen3vltechnicalreport}. Specifically, we perform physical-attribute rewriting to replace generic object names in captions with verified physical descriptions, ensuring the model learns robust physics-semantic alignments.

\begin{table*}[t]
\centering
\caption{Performance of leading MLLMs on \textbf{Physics Benchmarks} and our proposed \textbf{FysicsEval}.}
\label{tab:physics-bmk}
\setlength{\tabcolsep}{4pt} 
\renewcommand{\arraystretch}{1} 
\resizebox{\linewidth}{!}{%
\begin{tabular}{lcccccccc}
\toprule
\multirow{2}{*}{\textbf{Model}} & \multirow{2}{*}{\textbf{Size}} &
\multicolumn{3}{c}{\textbf{FysicsEval}} &
\multirow{2}{*}{\textbf{PhysBench}} &
\multirow{2}{*}{\textbf{PAI-Bench}} &
\multirow{2}{*}{\textbf{QuantiPhy}} &
\multirow{2}{*}{\textbf{PhysUniBench}} \\
\cmidrule(lr){3-5}
& & \textbf{Prediction} & \textbf{Reasoning} & \textbf{Understanding} & & & & \\
\midrule
GPT-5~\cite{openai_introducing_gpt5_2025} & - & \textbf{40.3} & \textbf{3.48} & \textbf{89.9} & \textbf{64.4} & \textbf{68.5} & \underline{32.6} & \underline{71.9} \\
Gemini-2.5-flash~\cite{comanici2025gemini2.5} & - & 19.8 & \underline{3.10} & \underline{89.4} & \underline{56.4} & \underline{57.6} & \textbf{48.6} & 65.5 \\
Claude-4.5-Haiku~\cite{anthropic2025claudeHaiku45} &-  & \underline{35.3} & 2.89 & 60.3 & 45.4 & 41.7 & 22.8 & \textbf{75.0} \\
\midrule
Qwen3-VL-8B-Instruct~\cite{bai2025qwen3vltechnicalreport} & 8B & 20.1 & \underline{2.65} & \underline{90.1} & \textbf{52.1} & \underline{55.4} & 33.7 & \underline{44.6} \\
InternVL3.5-8B~\cite{wang2025internvl3} & 8B & 21.7 & 2.53 & 80.7 & 31.4 & 40.3 & \underline{35.4} & 43.8 \\
Ovis2.5~\cite{lu2025ovis25} & 2B & 20.4 & 2.46 & 89.5 & 43.8 & 42.7 & 29.3 & 37.0 \\
SAIL-VL2~\cite{yin2025sailvl} & 2B & \underline{21.9} & 2.58 & 84.7 & 44.4 & 48.1 & 25.6 & 37.3 \\
Qwen2.5-Omni~\cite{xu2025qwen2.5omni} & 3B & 18.1 & 1.71 & 87.5 & 35.5 & 50.6 & 28.2 & 33.4 \\
\rowcolor{dino}  \textbf{OmniFysics (Ours)} & 3B & \textbf{32.6} & \textbf{3.22} & \textbf{94.7} & \underline{47.2} & \textbf{57.7} & \textbf{38.5} & \textbf{50.8} \\
\bottomrule
\end{tabular}%
}
\end{table*}
\begin{table*}[t]
\caption{Performance of OmniFysics on \textbf{Omni-modal and Video Understanding} Benchmarks compared to leading OLMs.}
\label{tab:omni-bmk}
\setlength{\tabcolsep}{12pt}
\renewcommand{\arraystretch}{1}
\resizebox{\linewidth}{!}{%
\begin{tabular}{lccccccc}
\toprule
\textbf{Model} & \textbf{Size} & \textbf{OmniBench} & \textbf{WorldSense} & \textbf{DailyOmni} & \textbf{FysicsWorld} & \textbf{Video-MME} & \textbf{Average} \\ \hline 
Qwen3-Omni-30B-A3B~\cite{xu2025qwen3omnitechnicalreport} & 30B & \textbf{58.41} & \textbf{52.01} & \textbf{75.80} & \textbf{67.14} & \textbf{70.5} & \textbf{64.77} \\
Qwen2.5-Omni~\cite{xu2025qwen2.5omni}  & 7B & {\ul 56.13} & 45.40 & 47.45 & {\ul 58.58} & 64.3 & 54.37 \\
OmniVinci~\cite{ye2025omnivinci}  & 7B & 46.47 & {\ul 48.23} & {\ul 66.50} & 55.52 & {\ul 68.2} & {\ul 56.98} \\
Unified-IO-2 XXL~\cite{lu2024unifiedio2} & 7B & 33.98 & 25.90 & 28.24 & 47.62 & 54.4 & 38.03 \\ \hline
Unified-IO-2 L~\cite{lu2024unifiedio2} & 1B & 27.06 & 23.30 & 27.40 & 45.34 & 45.2 & 33.66 \\
Unified-IO-2 XL~\cite{lu2024unifiedio2} & 3B & 38.00 & 24.70 & 28.30 & 47.62 & 46.8 & 37.08 \\
Qwen2.5-Omni~\cite{xu2025qwen2.5omni} & 3B & {\ul 45.18} & {\ul 44.45} & \textbf{40.52} & {\ul 51.49} & {\ul 62.0} & {\ul 48.73} \\
\rowcolor{dino}  \textbf{OmniFysics (Ours)} & 3B & \textbf{47.27} & \textbf{45.39} & {\ul 39.17} & \textbf{54.24} & \textbf{63.8} & \textbf{49.97} \\
\bottomrule
\end{tabular}
}
\end{table*}

\vspace{0.5pt}\noindent\textbf{Objective.}\quad
We train the DiT flow head via conditional flow matching. We define the probability path from noise $\mathbf{x}_0$ to data $\mathbf{x}_1$ as $\mathbf{x}_t = (1-t)\mathbf{x}_0+t\mathbf{x}_1$. The objective is:
\begin{small}%
\begin{equation}
\mathcal{L}_{\mathrm{img\_gen}}
= \mathbb{E}_{t,\mathbf{x}_0,\mathbf{x}_1}
\left\|v_{\eta}(\mathbf{x}_t,t,H)-(\mathbf{x}_1-\mathbf{x}_0)\right\|_2^2,
\label{eq:loss_img_gen}
\end{equation}
\end{small}%
where $\mathbf{x}_1$ is the VAE latent of the target image, $\mathbf{x}_0\sim\mathcal{N}(0,I)$, and $H$ is the multimodal conditioning context.

\subsection{Adaptive Reasoning}
\label{sec:adaptive_reason}
To balance efficiency with interactivity, we propose an intent-aware router (SA-IAR). This lightweight module processes text inputs $x$ to dynamically toggle the Flow Head with negligible latency.

\vspace{0.5pt}\noindent\textbf{Hybrid Feature Routing.}\quad
SA-IAR enhances boundary instruction discrimination by fusing ModernBERT~\cite{warner2025smarter} semantic embeddings $\mathbf{e}_{sem}$ with a hand-crafted syntactic vector $\mathbf{v}_{syn}$. To improve robustness against ambiguous inputs, $\mathbf{v}_{syn}$ is explicitly composed of four domain-specific indicators: (1) \textit{visual imperative flags} (\eg, ``simulate'', ``draw''), (2) \textit{cognitive verb flags} (\eg, ``explain'', ``define''), (3) \textit{causal query markers }(``why''), and (4) \textit{physical entity density}. The intent probability $s$ is computed via a gated network $G$:
\begin{small}%
\begin{equation}
s = \sigma(G([\mathbf{e}_{sem}, \mathbf{v}_{syn}])).
\label{eq:loss_adaptive_reason}
\end{equation}
\end{small}%
This integration of syntactic priors robustly aids semantic judgment, particularly in distinguishing visual generation requests from pure text inquiries.

\vspace{0.5pt}\noindent\textbf{Dynamic Execution \& Adversarial Training.}\quad
The system employs a branching strategy based on $s$: \textbf{Understanding Mode} ($s < \tau$): The Flow Head remains inactive. Low-entropy inputs (\eg, chitchat) trigger a direct response path to minimize latency. \textbf{Visual Generation Mode} ($s \ge \tau$): The system leverages the LLM's hidden states as conditional inputs to activate the Flow Head. To prevent syntactic overfitting, we utilize hard negative samples to ensure the generation module activates only when strictly necessary. These samples consist of requests containing high-density visual keywords but requiring textual outputs (\eg, ``\textit{Write a Python script to \textbf{simulate} a pendulum}'' or ``\textit{\textbf{Draw} a conclusion from the data}'').
\section{Experiments and Results}
\label{sec:experiments}

In this section, we systematically evaluate OmniFysics. First, we assess its omni-modal understanding across physical perception, image, video, and audio domains. This evaluation utilizes our proposed FysicsEval alongside other standard benchmarks. Next, we investigate the model's physics-aware generation capabilities. We compare OmniFysics against leading Expert and Omni Models. These experiments validate its superiority in synthesizing high-fidelity imagery with robust physical faithfulness.

\subsection{Physical Perception Evaluation}

We first evaluate leading MLLMs on physical understanding and reasoning across a diverse set of benchmarks, including the proposed \textbf{FysicsEval}, \textbf{PhysBench}~\cite{chow2025physbench}, \textbf{PAI-Bench}~\cite{zhou2025paibench}, \textbf{QuantiPhy}~\cite{puyin2025quantiphy}, and \textbf{PhysUniBench}~\cite{wang2025physunibench}. Then we compare OmniFysics against a comprehensive suite of state-of-the-art MLLMs. These are categorized into closed-source systems and open-source models spanning different parameter scales.

As presented in Table~\ref{tab:physics-bmk}, OmniFysics demonstrates exceptional performance on physical perception and reasoning benchmarks, not only significantly outperforming models of comparable scale but also exhibiting strong competitiveness against larger state-of-the-art systems. Specifically, within the sub-4B parameter category, our model consistently surpasses its open-source counterparts. On the proposed FysicsEval, OmniFysics achieves 32.6 in Prediction and 3.22 in Reasoning, nearly doubling the scores of its direct counterpart, Qwen2.5-Omni 3B (18.1 and 1.71, respectively). It also establishes a substantial lead on external benchmarks, such as securing 50.8 on PhysUniBench compared to Qwen2.5-Omni's 33.4.
Notably, OmniFysics displays a capability to transcend scaling laws in specific domains, outperforming larger open-source models like Qwen3-VL-8B-Instruct. More impressively, our lightweight architecture demonstrates formidable competitiveness against leading closed-source systems. It significantly surpasses Gemini-2.5-flash by a large margin in FysicsEval while maintaining a competitive edge in PAI-Bench. Furthermore, OmniFysics secures the second-highest score behind GPT-5 in FysicsEval Understanding (94.7). These results validate that our specialized physical data engines effectively bridge the gap between compact architectures and high-density physical knowledge, enabling the emergence of robust physical intelligence without the need for massive parameter scaling.

\subsection{General Multimodal Understanding}

To verify the model's versatility beyond specialized physical domains, we conduct a comprehensive evaluation on general multimodal benchmarks. This assessment spans three core categories: dynamic omni-modal and video understanding, static vision-language tasks, and audio-language analysis.
\begin{figure}[tp]
    \centering
    \includegraphics[width=\linewidth]{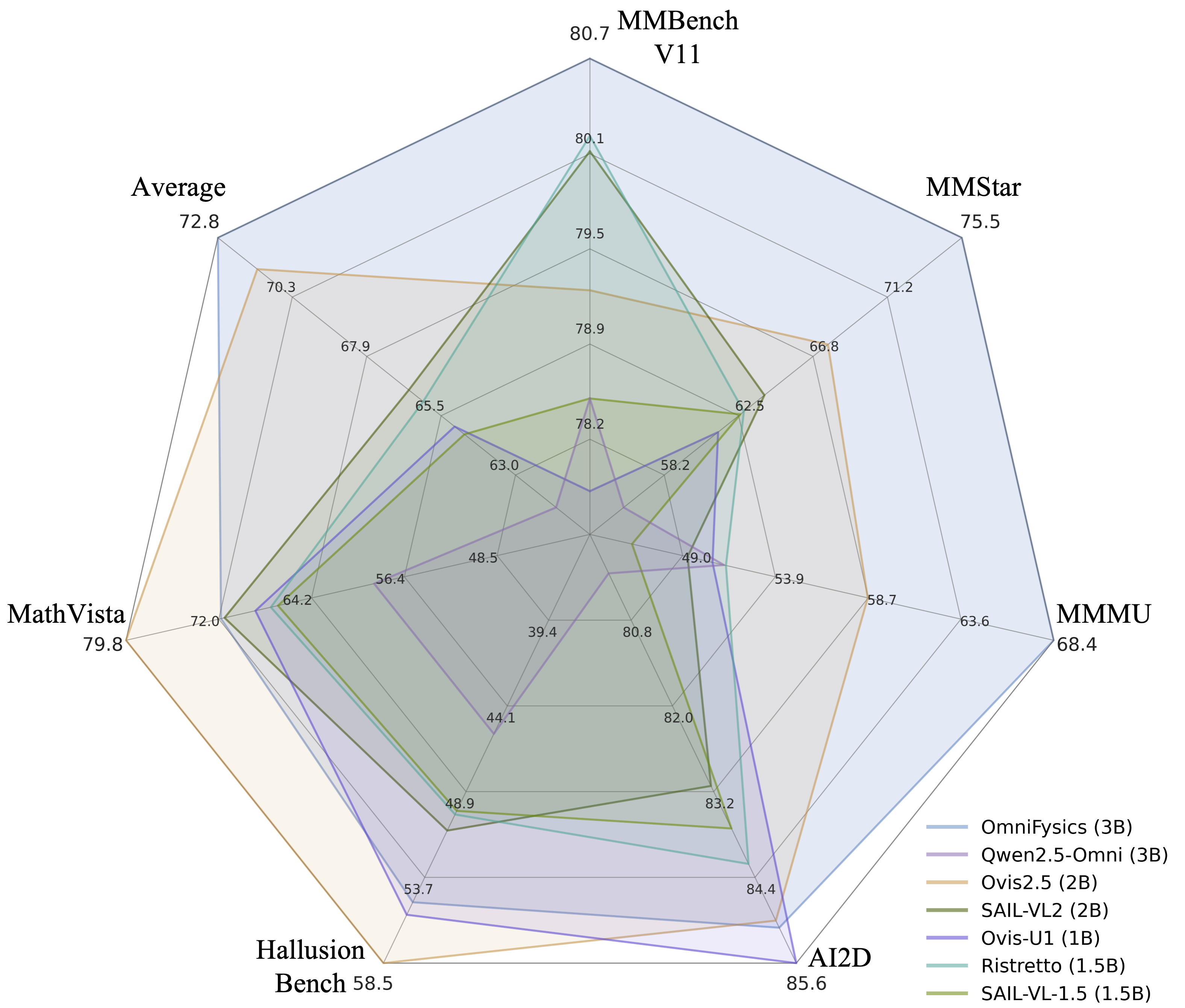}
    \caption{OmniFysics on Vision Understanding vs. Leading $<$4B MLLMs.}
    \label{fig:image_bmk}
\end{figure}

\begin{table}[thbp]
\centering
\captionsetup{width=\linewidth, justification=raggedright, singlelinecheck=false}
\caption{Performance on audio understanding and benchmarks: OmniFysics vs. leading MLLMs.}
\label{tab:audio-bmk}
\setlength{\tabcolsep}{4pt}
\renewcommand{\arraystretch}{1.1}
\resizebox{\linewidth}{!}{%
\begin{tabular}{lccccc}
\hline
\textbf{Model}             & \textbf{Size} & \textbf{\begin{tabular}[c]{@{}c@{}}MMAU\\ (ACC$\uparrow$)\end{tabular}} & \textbf{\begin{tabular}[c]{@{}c@{}}MMAR\\ (ACC$\uparrow$)\end{tabular}} & \textbf{\begin{tabular}[c]{@{}c@{}}Seed-TTS-zh\\ (WER$\downarrow$)\end{tabular}} & \textbf{\begin{tabular}[c]{@{}c@{}}Seed-TTS-en\\ (WER$\downarrow$)\end{tabular}} \\ \hline
\multicolumn{6}{c}{\textit{Size \textgreater 4B}}                                                                                                                                                                                                                                                                        \\ \hline
Qwen2.5-Omni~\cite{xu2025qwen2.5omni}               & 7B            & 71.50                                                         & 56.70                                                         & 1.70                                                                 & 2.72                                                                 \\
Qwen2-Audio-Instruct~\cite{chu2024qwen2audio}       & 7B            & 59.60                                                         & 33.33                                                         & -                                                                    & -                                                                    \\
Kimi-Audio~\cite{ding2025kimiaudio}                 & 7B            & 68.20                                                         & 38.79                                                         & -                                                                    & -                                                                    \\
Audio Reasoner~\cite{xie2025audioreasoner}             & 7B            & 67.70                                                         & 36.80                                                         & -                                                                    & -                                                                    \\
Baichuan-Omni-1.5~\cite{li2025baichuan}          & 7B            & 62.50                                                         & 40.70                                                         & 1.87                                                                 & 2.62                                                                 \\
GPT-4o Audio~\cite{gpt4o}               & -             & 62.50                                                         & 63.50                                                         & 1.41                                                                 & 2.19                                                                 \\
Gemini-2.5-Pro~\cite{comanici2025gemini2.5}             & -             & 71.60                                                         & 74.70                                                         & 1.35                                                                 & 1.98                                                                 \\ \hline
\multicolumn{6}{c}{\textit{Size \textless 4B}}                                                                                                                                                                                                                                                                           \\ \hline
Gemma 3n~\cite{team2025gemma3}                   & 2B            & 51.69                                                         & 37.80                                                         & -                                                                    & -                                                                    \\
Gemma 3n~\cite{team2025gemma3}                   & 4B            & 58.00                                                         & 40.10                                                         & -                                                                    & -                                                                    \\
Audio Flamingo 2~\cite{ghosh2025audioflamingo}           & 3B            & 62.40                                                         & 21.90                                                         & -                                                                    & -                                                                    \\
Qwen2.5-Omni~\cite{xu2025qwen2.5omni}              & 3B            & \textbf{66.30}                                                & {\ul 53.80}                                                   & {\ul 1.97}                                                           & {\ul 2.84}                                                           \\
\rowcolor{dino}  \textbf{OmniFysics (Ours)} & 3B            & {\ul 65.60}                                                   & \textbf{56.80}                                                & \textbf{1.45}                                                        & \textbf{2.33}                                                        \\ \hline
\end{tabular}
}
\end{table}

\vspace{0.5pt}\noindent\textbf{Omni \& Video Understanding.}\quad
We evaluate comprehensive understanding under omni-modal inputs, as well as temporal capabilities in video scenarios. Specifically, we utilize \textbf{OmniBench}~\cite{li2024omnibench}, \textbf{WorldSense}~\cite{hong2025worldsense}, \textbf{Daily-Omni}~\cite{zhou2025daily}, \textbf{FysicsWorld}~\cite{jiang2025fysicsworld}, and \textbf{Video-MME}~\cite{fu2025videomme}. We compare our model against Qwen2.5-Omni~\cite{xu2025qwen2.5omni}, and Unified-IO-2~\cite{lu2024unifiedio2}. As illustrated in Table~\ref{tab:omni-bmk}, OmniFysics demonstrates exceptional parameter efficiency in omni-modal and video understanding tasks. It outperforms its direct competitor, Qwen2.5-Omni 3B, on 4 out of the 5 evaluated benchmarks. Overall, it achieves the best performance with a leading average score of 49.97. We attribute this consistent superiority to FysicsOmniCap, which empowers the model with dense, physics-grounded instruction data, significantly enhancing its reasoning capabilities regarding fine-grained spatiotemporal dynamics and causality.

\vspace{0.5pt}\noindent\textbf{Vision-Language Understanding.}\quad
Subsequently, to verify the generalization performance in general visual scenarios, we quantitatively evaluated OmniFysics on six representative visual understanding benchmarks: \textbf{MMBench V1.1}~\cite{liu2024mmbench} and \textbf{MMStar}~\cite{mmstar} for comprehensive multimodal perception, \textbf{MMMU}~\cite{yue2024mmmu} and \textbf{MathVista}~\cite{lu2023mathvista} for advanced disciplinary knowledge and mathematical reasoning, as well as \textbf{HallusionBench}~\cite{guan2024hallusionbench} and \textbf{AI2D}~\cite{ai2d} for hallucination robustness and scientific diagram understanding. Specifically, we benchmark OmniFysics against six leading MLLMs with fewer than 4B parameters. As shown in Figure~\ref{fig:image_bmk}, OmniFysics achieves superior average performance (72.8\%) among sub-4B models, validating that our four-stage training preserves general perception. The significant gains on MMMU (reaching 68.4) and MMStar (reaching 75.5) align well with the capabilities developed through our Physical Logic Reasoning tasks. This paradigm compels the model to infer latent causal relationships rather than surface-level patterns, enhancing robustness in complex disciplinary reasoning. Conversely, the dip on MathVista reflects a trade-off: optimizing for physical causal chains creates a strong physical prior that may interfere with processing purely abstract mathematical diagrams.

\begin{figure*}[tp]
    \centering
    \includegraphics[width=0.9\linewidth]{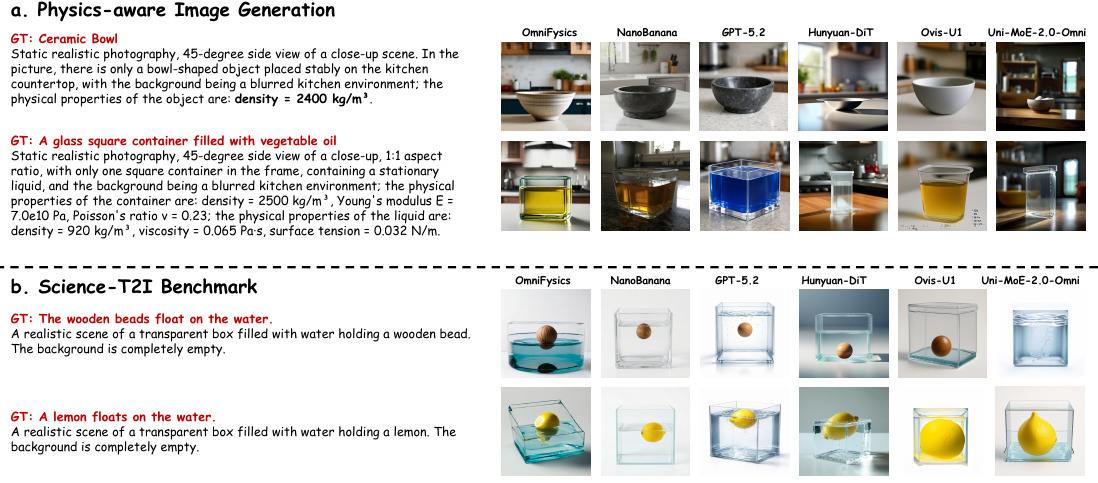}
    \caption{\textbf{Physics-aware Generation.} Mapping Physical Parameters to Faithful Materials and Simulating Accurate Scientific Phenomena.}
    \label{fig:image_gen}
    \vspace{-1em}
\end{figure*}


\vspace{0.5pt}\noindent\textbf{Audio Understanding \& Speech Generation.}\quad
Furthermore, we evaluate audio understanding capabilities on \textbf{MMAU}~\cite{sakshi2024mmau} and \textbf{MMAR}~\cite{ma2025mmar}, and assess speech generation proficiency using the Seed-TTS benchmarks for both Chinese (zh) and English (en). We benchmark against a comprehensive suite of baselines, ranging from large-scale proprietary systems to lightweight open-source models. As shown in Table~\ref{tab:audio-bmk}, OmniFysics demonstrates robust audio comprehension, achieving the accuracy of 65.60 on MMAU and 56.80 on MMAR. Notably, within the sub-4B parameter category, OmniFysics secures the highest MMAR score, outperforming Qwen2.5-Omni 3B (53.80) and significantly surpassing Gemma 3n 4B (40.10). With an average understanding score of 61.2, it remains highly competitive with the proprietary GPT-4o Audio (average 63.0) and exceeds the performance of multiple 7B models. 

In addition to comprehension, OmniFysics exhibits exceptional text-to-speech (TTS) capabilities. Evaluated by \textit{Word Error Rate} (WER, where lower is better), our model achieves a WER of 1.45 on Seed-TTS-zh and 2.33 on Seed-TTS-en. This performance establishes a new state-of-the-art among models under 4B, substantially reducing the WER compared to its direct counterpart, Qwen2.5-Omni 3B (1.97 and 2.84, respectively). Remarkably, our lightweight 3B architecture even outperforms larger 7B models such as Baichuan-Omni-1.5 and Qwen2.5-Omni 7B, while closely approaching the generation quality of leading proprietary systems like GPT-4o Audio (1.41/2.19) and Gemini-2.5-Pro (1.35/1.98). Overall, these results confirm that our architecture establishes highly competitive auditory understanding and speech generation without compromising the robust visual perception highlighted in previous sections.

\subsection{Physics-aware Generation}

\begin{table}[t!]
\centering
\captionsetup{width=\linewidth, justification=raggedright, singlelinecheck=false}
\caption{Performance of OmniFysics on \textbf{Image Generation Benchmarks} compared to leading Expert and Omni Models.}
\label{tab:image-gen-bmk}
\setlength{\tabcolsep}{4pt} 
\renewcommand{\arraystretch}{1.1}
\resizebox{\linewidth}{!}{%
\begin{tabular}{lccccr}
\toprule
\textbf{Model}           & \textbf{Size} & \textbf{GenEval} & \textbf{DPG-Bench} & \textbf{Science-T2I-S} \\ \hline
\multicolumn{5}{c}{\textit{Expert Models}} \\
\hline
SDv1.5~\cite{Rombach_2022_CVPR}                   & -             & 0.43             & 63.18              & 5.81 \\
Hunyuan-DiT~\cite{li2024hunyuandit}              & 2B            & 0.63             & 78.87              & 26.00 \\
Janus~\cite{wu2024janus}                    & 2B            & 0.61             & 79.68              & 30.78 \\
Ovis-U1~\cite{wang2025ovisu1}                  & 3B            & \textbf{0.89}             & \underline{83.72}              & \textbf{37.43} \\
OmniGen2~\cite{wu2025omnigen2}                 & 3B            & \underline{0.80}             & 83.57              & \underline{33.68} \\
Show-o2~\cite{xie2025showo2}                  & 2B            & 0.73             & \textbf{85.02}              & 25.76 \\ \hline
\multicolumn{5}{c}{\textit{Omni Models}}                                       \\ \hline
NExT-Omni~\cite{luo2025nextomni}        & 7B            & \textbf{0.85}             & \textbf{84.46}              & - \\
Uni-MoE-2.0-Omni~\cite{li2025unimoeomni} & 26B           & 0.61                & 77.41                  & \underline{33.61} \\
Ming-Lite-Omni~\cite{ai2025mingomni}   & 7B            & \underline{0.64}             & \underline{81.72}              & 27.35 \\
NExT-GPT~\cite{wu2024nextgpt}         & 7B            & 0.48                & 66.81                  & 19.87 \\
AnyGPT~\cite{zhan2024anygpt}           & 7B            & 0.52                & 74.44                  & 22.79 \\
Unified-IO-2 XL~\cite{lu2024unifiedio2}          & 3B            & 0.54                & 72.36                  & 9.88 \\ \hline
\rowcolor{dino}  \textbf{OmniFysics (Ours)}      & 3B            & 0.63               & 76.49                  & \textbf{38.02} \\ \bottomrule
\end{tabular}%
}
\vspace{-1.8em}
\end{table}

Finally, we assess the text-to-image generation capability of OmniFysics. We evaluate on \textbf{GenEval}~\cite{ghosh2023geneval}, \textbf{DPG-Bench}~\cite{dpg-bench}, and the physical-related subset of \textbf{Science-T2I-S}~\cite{li2025science}. We compare against two baseline categories: expert generation models and omni models with unified processing. As shown in Table~\ref{tab:image-gen-bmk}, OmniFysics achieves a score of 0.63 on GenEval. This performance substantially surpasses the comparable 3B omni-modal baseline Unified-IO-2 XL and yields results competitive with the expert model Hunyuan-DiT. This validates the efficacy of our progressive resolution upscaling strategy combined with the extensive training on 100M high-quality samples in establishing robust foundational image quality.

Crucially, OmniFysics demonstrates strong competitiveness on the physics-focused Science-T2I benchmark, outperforming the majority of 7B omni models. This physical property alignment is visually highlighted in Figure~\ref{fig:image_gen}. Specifically, as shown in Figure~\ref{fig:image_gen}a, while expert models generate high-fidelity objects, they frequently disregard explicit physical constraints (\eg, density $\rho$, Young's modulus $E$) in the prompt, defaulting to generic materials. In contrast, empowered by Physics-Grounded Refinement, OmniFysics accurately maps abstract physical parameters to visual materials—rendering ceramic-like textures for $\rho=2400 \text{kg/m}^3$ or rubber-like elasticity for low $E$. Furthermore, Figure~\ref{fig:image_gen}b further validates this physical faithfulness by showcasing its ability to accurately simulate natural scientific phenomena (\eg, buoyancy). This capability to translate physical laws into visual attributes shows physical faithfulness comparable to leading closed-source models.

\subsection{Ablation Study}
We conduct comprehensive ablation studies across multiple key dimensions to validate our core designs (Table~\ref{tab:ablation_study}).

\begin{figure*}[t!]
    \centering
    \includegraphics[width=\linewidth]{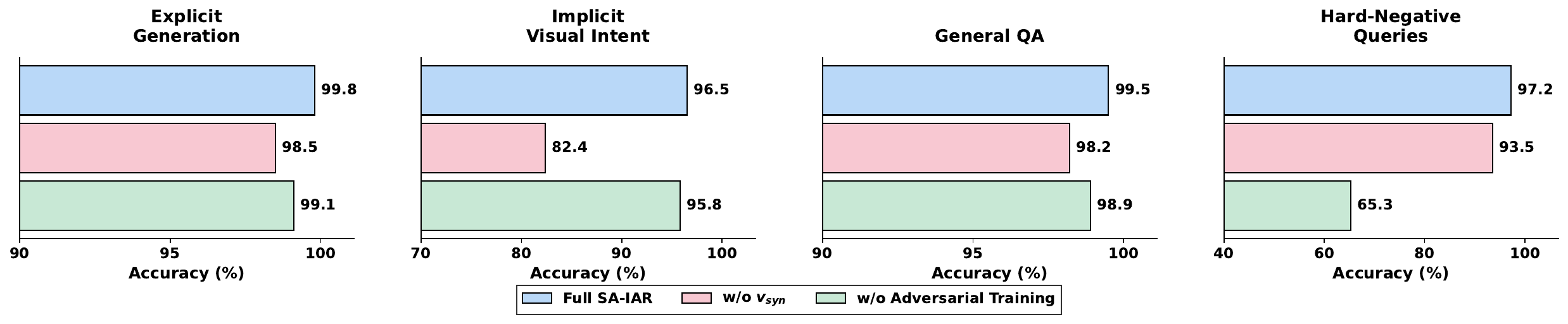}
    \caption{Routing accuracy of the intent-aware router (SA-IAR) and its ablated variants across four distinct interaction scenarios, demonstrating the specific impact of the syntactic vector ($\mathbf{v}_{syn}$) and adversarial training.}
    \label{fig:router_ablation}
    \vspace{-1.5em}
\end{figure*}

\vspace{0.5pt}\noindent\textbf{Impact of Physical Data Engines.}\quad
To manage computational costs, we construct a 5M proxy subset for data engine ablations. Crucially, variants omitting FysicsAny or FysicsOmniCap are padded with an equivalent volume of random same-modality data to ensure a fair comparison. Compared to the general baseline, injecting FysicsOmniCap significantly enhances dynamic tasks and cross-modal reasoning (MMAU +2.8, OmniBench +5.7). Conversely, FysicsAny markedly boosts static physical perception (MMMU +2.2, FysicsEval-Und. +3.4). Combining both yields the best overall performance (FysicsEval-Und. reaching 92.1), demonstrating that our data engines offer orthogonal and highly synergistic benefits.

\begin{table}[t!]
\centering
\caption{Ablation study across three scopes: physical data engines, data mixing strategies, and progressive training stages.}
\label{tab:ablation_study}
\resizebox{\linewidth}{!}{%
\begin{tabular}{l|cccc}
\toprule
\multicolumn{1}{c|}{\textbf{Components/}} & \textbf{Vision} & \textbf{Audio} & \textbf{Omni} & \textbf{Physics} \\
\multicolumn{1}{c|}{\textbf{Training Variants}} & MMMU & MMAU & OmniBench & FysicsEval-Und. \\
\midrule
\midrule
\multicolumn{5}{c}{Impact of Physical Data Engines (5M Subset)} \\
\midrule
General Baseline & 64.5 & 59.6 & 36.6 & 86.0 \\
\quad + FysicsOmniCap & 65.1 & 62.4 & 42.3 & 87.2 \\
\quad + FysicsAny & 66.7 & 59.5 & 37.1 & 89.4 \\
\midrule
\rowcolor{dino}  \textbf{Full Data (Ours)} & \textbf{67.1} & \textbf{62.4} & \textbf{42.5} & \textbf{92.1} \\
\midrule
\midrule
\multicolumn{5}{c}{Impact of Data Mixing \& Sampling Strategies (5M Subset)} \\
\midrule
Random Sampling & 66.3 & 61.5 & 41.2 & 89.8 \\
Heuristic Uniform & 65.1 & 62.0 & 42.1 & 88.6 \\
\midrule
\rowcolor{dino}  \textbf{Entropy-Guided (Ours)} & \textbf{67.1} & \textbf{62.4} & \textbf{42.5} & \textbf{92.1} \\
\midrule
\midrule
\multicolumn{5}{c}{Impact of Progressive Training Stages (Full 37M)} \\
\midrule
Stage 1: Modality-specific & 63.2 & 57.8 & 31.3 & 84.6 \\
Stage 2: Omni-modal Joint & \textbf{68.7} & 64.1 & 47.0 & \textbf{95.1} \\
\rowcolor{dino}  Stage 3: Audio Generation & 68.4 & \textbf{65.6} & \textbf{47.3} & 94.7 \\
\bottomrule
\end{tabular}
}
\vspace{-1.7em}
\end{table}

\vspace{0.5pt}\noindent\textbf{Impact of Data Mixing and Sampling Strategies.}\quad
To validate our Information Entropy Maximization strategy, we similarly compare it against two baseline strategies on a 5M proxy subset. \textbf{(1) Random Sampling}: Retains our optimal proportions (\eg, 48\% image, 16\% sound) but randomly selects samples instead of utilizing IDF-based filtering. \textbf{(2) Heuristic Uniform}: Applies IDF filtering but artificially forces an equal 25\% proportion across image, audio, video, and omni (\eg, audio-visual) modalities. Results indicate that Random Sampling degrades physical reasoning capabilities (FysicsEval-Und. drops by 2.3\%), highlighting the necessity of IDF scoring for isolating high-density physical concepts. Meanwhile, the Heuristic Uniform baseline underperforms because it ignores the varying information densities across different modalities. In contrast, our full Entropy-Guided approach optimally balances physical and semantic learning, achieving the highest overall performance.

\vspace{0.5pt}\noindent\textbf{Impact of Progressive Training Stages.}\quad
We further ablate our multi-stage training paradigm using the full dataset. The transition to Stage 2 (Omni-modal Joint Training) introduces the most substantial performance leaps across the board. Crucially, integrating audio generation in Stage 3 further enhances audio and omni-modal understanding (MMAU reaching 65.6, OmniBench reaching 47.3) while maintaining highly competitive performance in Vision and Physics (68.4 and 94.7, respectively). This confirms that our staged training smoothly harmonizes generation and perception without degrading previously acquired physical intelligence.

\vspace{0.5pt}\noindent\textbf{Impact of Adaptive Reasoning Router.}\quad
Finally, we ablate our intent-aware router (SA-IAR) across four fine-grained scenarios, each containing 1,000 prompts: Explicit Generation, Implicit Visual Intent, General QA, and Hard-Negative Queries. As visually summarized in Figure~\ref{fig:router_ablation}, our full SA-IAR  achieves exceptional routing accuracies across all four respective scenarios. Relying solely on ModernBERT semantic embeddings ($\mathbf{e}_{sem}$) and discarding the syntactic vector ($\mathbf{v}_{syn}$) causes the success rate on Implicit Visual Intent to plummet, as semantic embeddings alone struggle with nuanced linguistic structures. Furthermore, without adversarial training, the router suffers from severe syntactic overfitting. This drastically drops the accuracy on Hard-Negative Queries, which are text-only tasks containing misleading visual imperatives like simulate or draw. This confirms the necessity of fusing $\mathbf{e}_{sem}$ with $\mathbf{v}_{syn}$ and applying adversarial training for robust intent routing.
\section{Conclusion}
\label{sec:conclusion}
We present OmniFysics, a compact omni-modal model trained under web-scale supervision. Central to our approach is a physical data engine: FysicsAny constructs 4.7M physics-compliant instruction-image pairs through hierarchical retrieval and physics-law validation, while FysicsOmniCap generates 872K high-fidelity video-instruction pairs via audio-visual consistency filtering. Additionally, we introduced FysicsEval, a benchmark with 3,854 samples for comprehensively evaluating physical attribute prediction and causal reasoning. Architecturally, OmniFysics integrates flow-matching image generation with an intent-aware router to enable efficient interaction. Overall, results indicate that this physical data engine consistently improves model performance on physics-oriented understanding tasks. Future work will extend to long-horizon generation and embodied interaction simulation.

\bibliographystyle{IEEEtran}
\bibliography{reference}

\newpage  

\section{Biography Section}
 

\vspace{-33pt}

\begin{IEEEbiographynophoto}{Minghao Han}
received the B.S. degree in Automation from Shandong University, Jinan, China, in 2022. He is currently pursuing a doctoral degree at the College of Intelligent Robotics and Advanced Manufacturing, Fudan University, Shanghai. His research interests include omni-multimodal large language model, vision-language models, and computational pathology.
\end{IEEEbiographynophoto}

\vspace{-25pt}

\begin{IEEEbiographynophoto}{Dingkang Yang}
received the B.E. degree (with honors) in Communication Engineering from the joint training program of Yunnan University and the Chinese People's Armed Police (PAP), Kunming, China, in 2020, and the Ph.D. degree in Computer Science from Fudan University, Shanghai, China, in 2025. His research interests include multimodal learning, generative AI, and embodied AI. Dr. Yang has published multiple papers as the first author at the reputable journals and top international conferences, such as IEEE TPAMI, TCSVT, Information Fusion, NeurIPS, CVPR, ICCV, and ECCV. He has served as Special Session Co-Chair/Guiding Committee Expert for IEEE ICME 2026 and ICIP 2026, Guest Editor for the Journal of Real-time Image Processing, and TPC member for top-tier conferences including CVPR, ICCV, ICML, NeurIPS, and AAAI.
\end{IEEEbiographynophoto}

\vspace{-25pt}

\begin{IEEEbiographynophoto}{Yue Jiang}
received the B.E. degree in Software Engineering from Guangdong University of Foreign Studies, Guangzhou, China, in 2022. He is currently pursuing his Master's degree at the College of Intelligent Robotics and Advanced Manufacturing, Fudan University, Shanghai, China. His research interests include multimodal learning and trustworthy AI.
\end{IEEEbiographynophoto}

\vspace{-25pt}

\begin{IEEEbiographynophoto}{Yizhou Liu}
received his B.E. degree from the School of Computer Science and Technology, Harbin Institute of Technology, China, in 2025. He is currently pursuing the Ph.D. degree at the College of Intelligent Robotics and Advanced Manufacturing, at Fudan University, Shanghai, China. Meanwhile, he is working as a research intern at Fysics AI. His research interests mainly include omni-multimodal large language models, 3D reconstruction and reinforcement learning.
\end{IEEEbiographynophoto}

\vspace{-25pt}

\begin{IEEEbiographynophoto}{Peng Zhai}
received the Ph.D. degree in technology for computer applications from Fudan University, Shanghai, China, in 2022. He is currently a Research Fellow with the Institute of AI and Robotics, Fudan University, Shanghai, China. His research interests include Embodied intelligence, Robotics, and Reinforcement Learning.
\end{IEEEbiographynophoto}

\vspace{-25pt}

\begin{IEEEbiographynophoto}{Lihua Zhang}
received the Ph.D. degree from the Department of Automation, Tsinghua University, Beijing, China, in 2000. He is currently a Professor at the Academy for Engineering and Technology, Fudan University, Shanghai, China. In recent years, he has participated in a number of national science and technology research and development projects as a project leader and sub-project. His current research interests are in artificial intelligence and its applications, including machine intuition, computer vision and intelligent perception, virtual reality and digital twinning, intelligent robotics and unmanned systems, intelligent computing and intelligent chips, intelligent healthcare, intelligent connected vehicles, etc.
\end{IEEEbiographynophoto}



\vfill

\end{document}